\documentclass[review, 12pt, a4paper]{elsarticle}

\usepackage{amssymb} 
\usepackage{bm} 
\usepackage{hyperref}
\journal{SoftwareX}

\begin{document}
\renewcommand{\labelenumii}{\arabic{enumi}.\arabic{enumii}}

\begin{frontmatter}

\title{\textbf{\textsc{gamma\_flow}}: \textbf{G}uided \textbf{A}nalysis of \textbf{M}ulti-label spectra by \textbf{Ma}trix \textbf{F}actorization for \textbf{L}ightweight \textbf{O}perational \textbf{W}orkflows}

\author[ki-lab]{Viola Rädle \corref{cor1}}
\author[ki-lab]{Tilman Hartwig}
\author[ki-lab]{Benjamin Oesen}
\author[bfs]{Emily Alice Kröger}
\author[bfs]{Julius Vogt}
\author[bfs]{Eike Gericke}
\author[bfs]{Martin Baron}

\address[ki-lab]{Application Lab for AI and Big Data, German Environmental Agency, Leipzig, Germany}
\address[bfs]{Federal Office for Radiation Protection, Berlin, Germany}

\cortext[cor1]{Corresponding author: raedle.htwk - AT - web.de}

\begin{abstract}
\textsc{gamma\_flow} is an open-source Python package for real-time analysis of spectral data. It supports classification, denoising, decomposition, and outlier detection of both single- and multi-component spectra. Instead of relying on large, computationally intensive models, it employs a supervised approach to non-negative matrix factorization (NMF) for dimensionality reduction. This ensures a fast, efficient, and adaptable analysis while reducing computational costs. \textsc{gamma\_flow} achieves classification accuracies above 90\% and enables reliable automated spectral interpretation. Originally developed for gamma-ray spectra, it is applicable to any type of one-dimensional spectral data. As an open and flexible alternative to proprietary software, it supports various applications in research and industry.
\end{abstract}


\end{frontmatter}

\section*{Metadata}

\begin{table}[!ht]
\begin{tabular}{|l|p{6.5cm}|p{6.5cm}|}
\hline
\textbf{Nr.} & \textbf{Code metadata description} & \textbf{Metadata} \\
\hline
C1 & Current code version & 0.9.0 \\
\hline
C2 & Permanent link to code/repository used for this code version & \url{https://github.com/Schlavioner/gamma_flow} \\
\hline
C3  & Permanent link to Reproducible Capsule & - \\
\hline
C4 & Legal Code License & BSD 3-Clause "New" or "Revised" License \\
\hline
C5 & Code versioning system used & git\\
\hline
C6 & Software code languages, tools, and services used & Python \\
\hline
C7 & Compilation requirements, operating environments \& dependencies & Python $\ge$ 3.12, matplotlib, numpy, pandas, scikit\_learn, scipy, seaborn, openpyxl \\
\hline
C8 & If available Link to developer documentation/manual & \url{https://github.com/Schlavioner/gamma_flow/blob/main/README.md}, 
\url{https://github.com/Schlavioner/gamma_flow/tree/main/documentation/HTML-documentation}\\
\hline
C9 & Support email for questions & raedle.htwk - AT - web.de\\
\hline
\end{tabular}
\caption{Code metadata (mandatory)}
\label{codeMetadata} 
\end{table}

\begin{table}[!ht]
\begin{tabular}{|l|p{6.5cm}|p{6.5cm}|}
\hline
\textbf{Nr.} & \textbf{(Executable) software metadata description} & \textbf{Please fill in this column} \\
\hline
S1 & Current software version & 0.9.0 \\
\hline
S2 & Permanent link to executables of this version & \url{https://github.com/Schlavioner/gamma_flow} \\
\hline
S3 & Permanent link to Reproducible Capsule & - \\
\hline
S4 & Legal Software License & BSD 3-Clause "New" or "Revised" License \\
\hline
S5 & Computing platforms/Operating Systems & Linux, Microsoft Windows\\
\hline
S6 & Installation requirements \& dependencies & Python $\ge$ 3.12, matplotlib, numpy, pandas, scikit\_learn, scipy, seaborn, openpyxl \\
\hline
S7 & If available, link to user manual - if formally published include a reference to the publication in the reference list & \url{https://github.com/Schlavioner/gamma_flow/blob/main/README.md} \\
\hline
S8 & Support email for questions & raedle.htwk - AT - web.de\\
\hline
\end{tabular}
\caption{Software metadata (optional)}
\label{executabelMetadata} 
\end{table}

\section{Motivation and significance}

Most radioactive sources can be identified by measuring their emitted radiation (X-rays and gamma rays), and visualizing them as a spectrum. In nuclear security applications, the resulting gamma spectra have to be analyzed in real-time as immediate reaction and decision making may be required. 
However, the manual recognition of isotopes present in a spectrum constitutes a 
strenuous, error-prone task that depends on expert knowledge. Hence, this raises 
the need for algorithms assisting in the initial categorization and recognizability 
of measured gamma spectra. 
The delineated use case brings along several requirements: 
\begin{itemize}
\item As mobile, room-temperature detectors are often deployed in nuclear security applications, the produced spectra typically exhibit a rather low energy resolution. In addition, a high temporal resolution is required (usually around one spectrum per second), leading to a low acquisition time and a low signal-to-noise ratio. Hence, the model must be robust and be able to handle noisy data.
\item For some radioactive sources, acquisition of training spectra may be challenging. Instead, spectra of those isotopes are simulated using Monte Carlo N-Particle (MCNP) code \cite{Kulesza2022}. In this process, energy deposition in a detector material is simulated, yielding spectra that can be used for model training. However, as not all physical effects are included, simulated spectra and measured spectra from real-world sources may differ, which is a constraint for model performance. On this account, preliminary data exploration is crucial to assess their similarity and to evaluate potential data limitations.
\item Lastly, not only the correct classification of single-label test spectra (stemming from one isotope) is necessary, but also the decomposition of linear combinations of various isotopes (multi-label spectra). Hence, classification approaches like k-nearest-neighbours that solely depend on the similarity between training and test spectra are not applicable. 
\end{itemize} 
This paper presents \textsc{gamma\_flow}, a Python package designed for the real-time analysis of one-dimensional spectra. It was developed in response to the practical challenges described above and supports the following tasks:
\begin{itemize}
\item classification of test spectra to identify their constituents, 
\item denoising to enhance visibility and reduce measurement noise,
\item outlier detection to evaluate the model's applicability to unknown spectra. 
\end{itemize}

Originally developed for gamma spectroscopy, \textsc{gamma\_flow} is applicable to various domains including material science \cite{Zatsepin2021a}, chemistry \cite{Roessler2018}, and environmental monitoring \cite{Qi2024}. It facilitates automated analysis in settings where fast interpretation of spectral data is essential. By integrating supervised dimensionality reduction with classical signal processing techniques, it extends prior works such as those of Bilton et al. \cite{Bilton2019} or Bandstra et al. \cite{Bandstra2020}, contributing to reproducible and interpretable spectral analysis pipelines.

\section{Software description}
\label{sec:software_description}

\textsc{gamma\_flow} is based on a dimensionality reduction model that constitutes a supervised approach to non-negative matrix factorization (NMF). More explicitly, the spectral data matrix is decomposed into the product of two low-rank matrices denoted as the scores (spectral data in latent space) and the loadings (transformation matrix or latent components). The loadings matrix is predefined and consists of the mean spectra of the training isotopes. Hence, by design, the scores axes correspond to the share of an isotope in a spectrum, resulting in an interpretable latent space. As a result, the classification of a test spectrum can be read directly from its (normalized) scores. In particular, shares of individual isotopes in a multi-label spectrum can be identified. This leads to an explainable quantitative prediction of the spectral constituents. 

For spectral denoising, the scores are transformed back into spectral space by applying the inverse model. This inverse transformation rids the test spectrum of noise and results in a smooth, easily recognizable denoised spectrum. 

If a test spectrum of an isotope is unknown to the model (i.e., this isotope was not included in model training), it can still be projected into latent space. However, when the latent space information (scores) are decompressed, the resulting denoised spectrum does not resemble the original spectrum any more. Some original features may not be captured while new peaks may have been fabricated. This can be quantified by calculating the cosine similarity between the original and the denoised spectrum, which can serve as an indicator of a test spectrum 
to be an outlier.

\subsection{Software architecture}
\begin{figure}[ht]
\includegraphics[width=\textwidth]{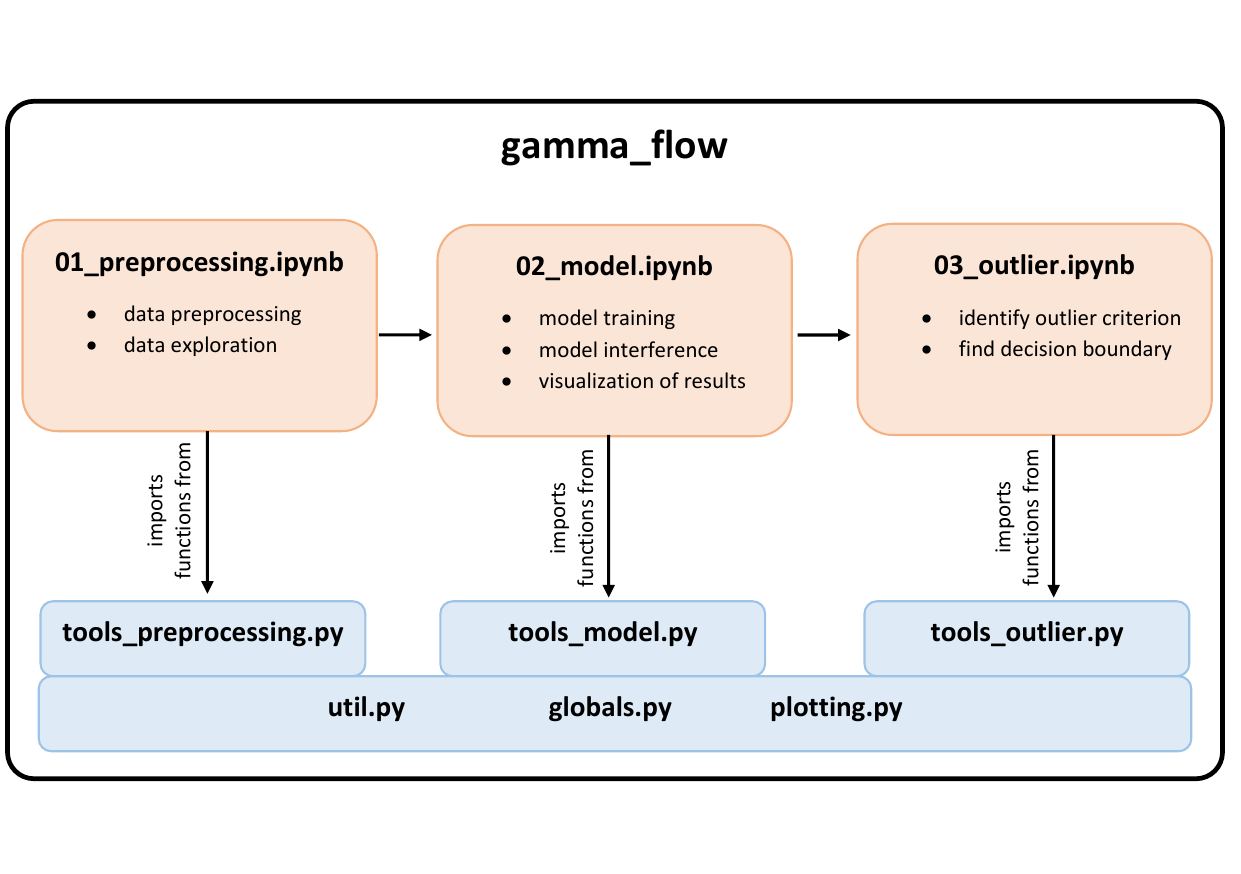}
\caption{Software architecture of \textsc{gamma\_flow}: The jupyter notebooks \texttt{01\_preprocessing.ipynb}, \texttt{02\_model.ipynb} and \texttt{03\_outlier.ipynb} are executed sequentially, using functions from different Python files. }
\label{fig:software_architecture}
\end{figure}

\textsc{gamma\_flow} consists of three jupyter notebooks that are executed consecutively, as depicted in Figure \ref{fig:software_architecture}. Each imports functions from a designated Python file, e.g. all functions called in \texttt{02\_model.ipynb} are found in \texttt{tools\_model.py}. In addition, the Python files \texttt{util.py}, \texttt{globals.py} and \texttt{plotting.py} provide elementary functions for all three notebooks. 

\subsection{Software functionalities}

In this section, the functionality of the software is outlined, with an emphasis on the mathematical structure of the model. To this end, the procedures realized in the three jupyter notebooks \texttt{01\_preprocessing.ipynb}, \texttt{02\_model.ipynb} and \texttt{03\_outlier.ipynb} are delineated. In the software repository, a sample dataset of gamma spectra with exemplary results is provided to give users a first insight into the software. 

\subsubsection{Preprocessing and data exploration}
The notebook \texttt{01\_preprocessing.ipynb} harmonizes spectral data and provides a framework of visualizations for data exploration. All functions called in this notebook are found in \texttt{tools\_preprocessing.py}. 

During \textbf{preprocessing}, the following steps are performed:
\begin{itemize}
\item Spectral data files are converted from .xlsm/.spe data to .npy format and saved.
\item Spectra of different energy calibrations are rebinned to a standard energy calibration.
\item Spectral data are aggregated by label classes and detectors. Thus, it is possible to 
 collect data from different files and formats. 
\item Optional: The spectra per isotope are limited to a maximum number (for class balance).
\item The preprocessed spectra are saved as .npy files.
\end{itemize}

\textbf{Data exploration} involves the following visualizations: 
\begin{itemize}
\item For each label class (e.g. for each isotope), the mean spectra are calculated detector-wise and compared quantitatively by the cosine similarity. 
\item For each label class, example spectra are chosen randomly and plotted to provide an overview over the data. 
\item The cosine similarity is calculated and visualized as a matrix for all label classes and detectors.
\end{itemize}
This helps to assess whether the model can handle spectra from different detectors.

\subsubsection{Model training and testing}
\label{sec:model_training_testing}
The notebook \texttt{02\_model.ipynb} trains and tests a dimensionality reduction model that allows for denoising, classification and outlier detection of test spectra. All functions called in this notebook are found in \texttt{tools\_model.py}. \\

\textbf{The model} presented in this paper performs a matrix decomposition of spectral data to achieve dimensionality reduction. More precisely, the original spectra matrix $\bm{X} \in \mathbb{R}_{\ge 0}^{n \times m}$, with $n$ spectra and $m$ channels, is reconstructed by two low-rank matrices: 
$$ \bm{X} \approx \bm{S} \bm{L}^\mathrm{T} $$
where $\bm{S} \in \mathbb{R}_{\ge 0}^{n \times k}$ is the scores matrix and $\bm{L} \in \mathbb{R}_{\ge 0}^{m \times k}$ the loadings matrix, with $k = k_\mathrm{isotopes}$ latent components. \\ 

\begin{figure}
\includegraphics[width=\textwidth]{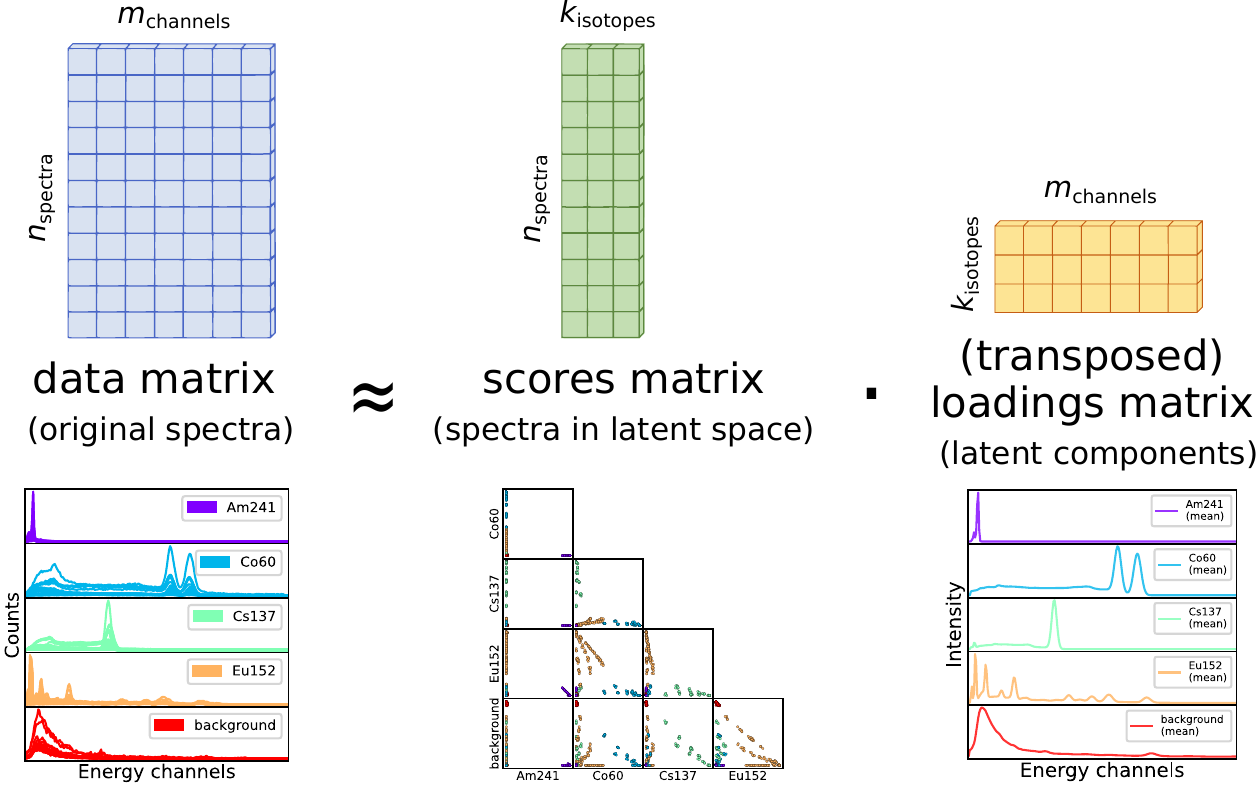}
\caption{Decomposition of spectral data into scores and loadings, illustrated for five different isotopes. The loadings matrix contains the latent components, corresponding to the mean spectrum of each isotope. The scores matrix provides the representation of each spectrum in the five-dimensional latent space. When normalized, the scores indicate the predicted contribution of each isotope to the measured spectrum.}
\label{fig:matrix_decomposition}
\end{figure}

As shown in Figure~\ref{fig:matrix_decomposition}, each spectrum is projected into a low-dimensional space spanned by the isotope-specific mean spectra. This corresponds to a supervised variant of Non-negative Matrix Factorization (NMF) \cite{Shreeves2020, Bilton2019}, where the loadings matrix $\bm{L}$ is predefined instead of being learned. The latent components thus retain physical interpretability, each representing the average spectral signature of one isotope.

To compute the scores $\bm{S}$, a non-negative least squares (NNLS) problem is solved for each spectrum $\bm{x}_\mathrm{i}$ (row vector of $\bm{X}$):
$$\bm{s}_\mathrm{i} = \mathrm{argmin}_{\bm{s} \ge 0} ||\bm{x}_\mathrm{i} - \bm{s} \bm{L}^\mathrm{T}||_2^2$$ 

This constrained optimization is carried out using the \texttt{scipy.optimize.nnls} implementation \cite{SciPy} based on the Lawson-Hanson active-set algorithm \cite{Lawson1995a}, which solves a sequence of unconstrained least squares problems by iteratively updating an active set of non-negativity constraints. The algorithm is guaranteed to converge to the global minimum of the convex problem for each spectrum and ensures strict non-negativity of the resulting coefficients. 

This enables an interpretable \textbf{spectral decomposition}, where the normalized scores vector $\bm{s}_\mathrm{i.norm} = \bm{s}_\mathrm{i} / \sum \bm{s}_\mathrm{i}$ directly reveals the relative contribution of each isotope. \textbf{Denoised spectra}, on the other hand, are computed by transforming the (non-normalized) scores back into spectral space by multiplication with the loadings matrix: $\bm{x}_\mathrm{i} = \bm{s}_\mathrm{i} \bm{L}^\mathrm{T}$. This removes noise orthogonal to the latent space. \\

To assess \textbf{model performance}, the model is trained using spectral data from the specified detectors \texttt{dets\_tr} and isotopes \texttt{isotopes\_tr} and inferenced (i.e., scores are calculated) on three different test datasets:
\begin{enumerate}
\item validation data/holdout data from same detector as used in training (simulated data: each spectrum including only one isotope or pure background)
\item test data from different detector (measured single-label data: each spectrum including one isotope and background)
\item multi-label test data from different detector (measured multi-label data: each spectrum including multiple isotopes and background)
\end{enumerate}

For all test datasets, spectra are classified and denoised. The results are visualized as 
\begin{itemize}
\item confusion matrix 
\item misclassified spectra 
\item denoised example spectrum 
\item misclassification statistics 
\item scores as scatter matrix 
\item mean scores as bar plot
\end{itemize}
 
This helps to assess model performance with respect to classification and denoising. 

\subsubsection{Outlier analysis}
\label{sec:outlier_analysis}
The notebook \texttt{03\_outlier.ipynb} provides an exploratory approach to outlier detection, i.e., to identify spectra from isotopes that were not used in model training. All functions called in this notebook are found in \texttt{tools\_outlier.py}. 

To simulate outlier spectra, a mock dataset is generated by training a model after removing one specific isotope. The trained model is then inferenced on spectra of this unknown isotope to investigate its behaviour with outliers. 
First, the resulting latent space distribution and further metadata are analyzed to distinguish known from unknown spectra. Using a decision tree, the most informative feature for this task is identified based on permutation feature importance, i.e., by measuring how much the performance drops when feature values are randomly shuffled \cite{scikit-learn2025}.
Next, a decision boundary is derived for this feature, by \\
a) using the condition of the first split in the decision tree, as depicted in \texttt{03\_outlier.ipynb}\\
b) fitting a logistic regression (sigmoid function) to the outlier data, with the most informative feature as x and outlier score as y (by setting known spectra to 0 and outliers to 1). The decision boundary can be read from the fit parameters. \\ 
c) setting a manual threshold by considering accuracy, precision and recall of outlier identification. This is done by plotting those quantities against the most informative feature and visually evaluating their optimal trade-off for this task. Further metrics, like True Positive Rate or False Positive Rate, can be calculated in the corresponding notebook. \\
The derived decision boundary can then be implemented in the measurement pipeline by the user. \\
The excellent model performance on the noisy gamma dataset at hand proves the model's applicability to noisy spectra. In addition, we have verified that adding Gaussian noise to the spectra has only a minor effect on classification and denoising results. In the context of outlier detection, higher ratios of noise result in slightly diminished cosine similarities, shifting the decision boundary towards smaller values. \\

Apart from the jupyter notebooks and python files described above, the project includes the following python files: 
\begin{itemize}
\item \texttt{globals.py}: global variables 
\item \texttt{plotting.py}: all visualizations and plotting routines 
\item \texttt{util.py}: basic functions that are used by all notebooks
\end{itemize}

\section{Illustrative example}

The major functions of \textsc{gamma\_flow} include classification, denoising, decomposition, and outlier detection of spectra. In this section, these capabilities are demonstrated using an illustrative example. 

A model is trained based on simulated spectra of the isotopes Americium-241 (\textsuperscript{241}Am), Cobalt-60 (\textsuperscript{60}Co), Caesium-137 (\textsuperscript{137}Cs), Europium-152 (\textsuperscript{152}Eu), all free of background radiation, as well as pure measured background spectra. 

The model is then inferenced on various test datasets to assess its performance in different scenarios (for more detail, see Section~\ref{sec:model_training_testing}). The following example focuses on a measured test dataset where spectra contain either a single isotope combined with background or pure background only. No background subtraction was applied.

\begin{figure}
\centering
\includegraphics[width=0.7\textwidth]{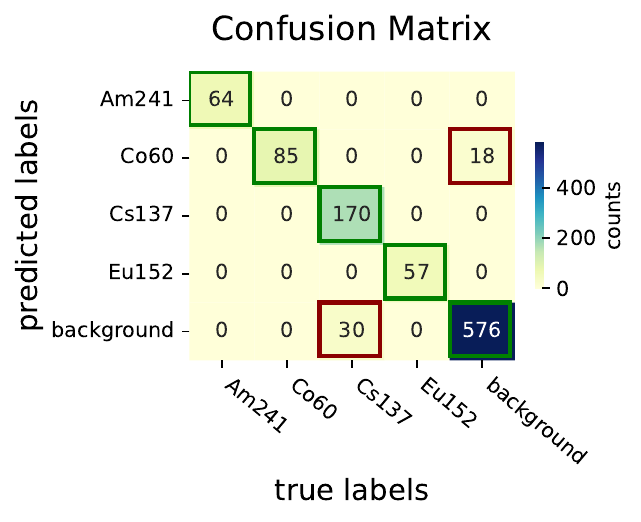}
\caption{Confusion matrix of single-label classification, with an accuracy of 94.8\%. While the model was trained on simulated spectra, measured spectra were used as test data. Misclassifications only occur between an isotope and background, not between different isotopes.}
\label{fig:confusion_matrix}
\end{figure}

The \textbf{classification} results for this test dataset can be seen in Figure \ref{fig:confusion_matrix}, where the predicted labels are plotted against the ground truth as a confusion matrix. Correct classifications are highlighted with green borders while misclassifications are framed in red. Despite the difference between training and test spectra (different detector and presence of background in test data), an overall accuracy of 94.8\% is achieved. Notably, all misclassifications involve confusion between an isotope and pure background — never between two isotopes. Such cases typically occur when the isotope signal is near the detection threshold and the model's prediction deviates from the manually assigned label. Please refer to the software documentation in the code repository for further visualizations on model training and classification results, such as the loadings, the scores (visualized as scatter matrix), the mean predicted scores by isotope, misclassified example spectra and misclassification statistics. \\

\begin{figure}
\includegraphics[width=\textwidth]{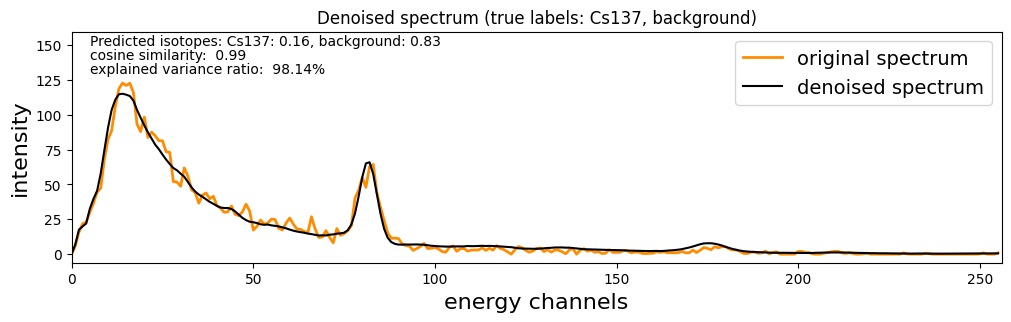}
\caption{Comparison between the original and the reconstructed denoised spectrum of \textsuperscript{137}Cs and background. The denoised spectrum is smoother, leading to an easy and fast interpretability.}
\label{fig:denoised_spectrum}
\end{figure}

As described in Section \ref{sec:model_training_testing}, \textbf{denoised spectra} are obtained via inverse transformation using the predefined loadings. The resulting reconstructed spectra are significantly smoother, facilitating the distinction between isotope and random features. A denoised example spectrum of \textsuperscript{137}Cs and background is depicted in Figure \ref{fig:denoised_spectrum}. As can be seen in the annotations, the predictions are correct, with an additional information on the contribution of the components: The model predicts 16\% share of \textsuperscript{137}Cs and 83\% background. The reconstruction of the original spectrum by the denoised spectrum is quantified by their cosine similarity of 0.99 and the explained variance ratio of 98.14\%, which both indicate a successful denoising process. \\

\begin{figure}
\includegraphics[width=\textwidth]{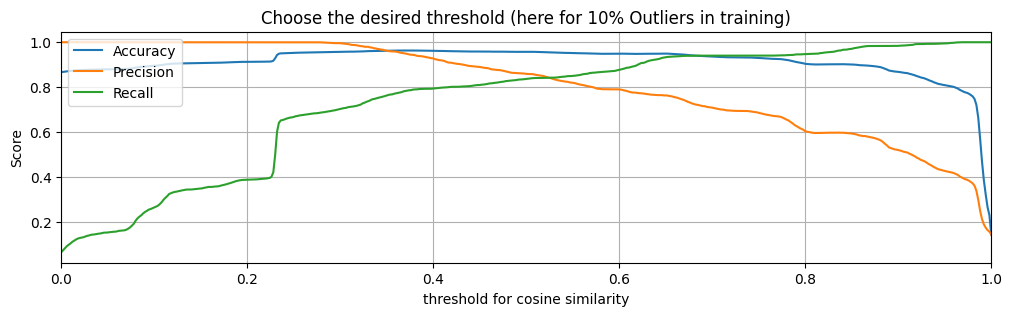}
\caption{For outlier detection, the decision boundary can be derived based on accuracy, precision and recall of the outliers dataset. }
\label{fig:outlier}
\end{figure}

The options for \textbf{outlier detection} are analyzed as described in Section \ref{sec:outlier_analysis}. After the most informative feature is determined (in our case the cosine similarity between original and denoised spectrum) a decision boundary is derived. This can be done manually by considering accuracy, precision and recall, as visualized in Figure \ref{fig:outlier}. Those quantities yield different information on the separation of known and unknown spectra:
\begin{itemize}
\item \textbf{Accuracy} quantifies the correct classifications as known and unknown spectra: $\frac{\mathrm{True\;Positives + True\;Negatives}}{\mathrm{True\;Positives + False\;Positives + True\;Negatives + False\;Negatives}}$
\item \textbf{Precision} quantifies how many of the predicted outliers are actually unknown spectra: $\frac{\mathrm{True\;Positives}}{\mathrm{True\;Positives + False\;Positives}}$
\item \textbf{Recall} quantifies how many of the unknown spectra are found: \\$\frac{\mathrm{True\;Positives}}{\mathrm{True\;Positives + False\;Negatives}}$
\end{itemize}
In this example, a threshold between 0.5 and 0.9 would be an adequate choice. 

\section{Impact and Discussion}

\subsection{Applications}
In many research fields, spectral measurements help to assess material properties. 
In this context, an area of interest for many researchers is the classification (automated labelling) of the measured spectra. Proprietary spectral analysis software, however, are often limited in their functionality and adaptability \cite{Lam2011, Nasereddin2023}. In addition, the underlying mechanisms are usually not revealed and may act as a black-box system to the user \cite{ElAmri2022}. On top of that, a spectral comparison is typically only possible for spectra of pure probes \cite{Cowger2021}. However, there may be a need to decompound multi-label spectra (linear combinations of different substances) and identify their constituents. 

\textsc{gamma\_flow} closes this gap and provides a lean and performant model. Training and inference do not require special hardware or extensive computational power. This allows real-time application on ordinary laboratory computers and easy implementation into the measurement routine. 

The provided example dataset contains gamma spectra of several measured and simulated isotopes as well as pure background spectra. While this package was developed in need of an analysis tool for gamma spectra, it is suitable for any one-dimensional spectra. Exemplary applications encompass
\begin{itemize}
\item \textbf{Infrared spectroscopy} for the assessment of the polymer composition of 
microplastics in water \cite{Ferreiro2023, Whiting2022}
\item \textbf{mass spectrometry} for protein identification in snake venom 
\cite{Zelanis2019, Yasemin2021}
\item \textbf{Raman spectroscopy} for analysis of complex pharmaceutical mixtures and detection of dilution products like lactose \cite{Fu2021}
\item \textbf{UV-Vis spectroscopy} for detection of pesticides in surface waters \cite{Guo2020, Qi2024}
\item \textbf{stellar spectroscopy} to infer the chemical composition of stars \cite{Gray2021}
\end{itemize}

\subsection{Comparison to Related Work}

Clustering and classification of spectral data have been explored through various NMF-based methods and have proven to excel at grouping unlabelled spectra \cite{Mirzal2022}. To target robustness or interpretability, Robust and Sparse NMF \cite{Zhang2011, Chen2024, Qu2024, Mella2022} mitigate noise and outliers through explicit error modelling or sparsity regularization. Meanwhile, Graph-regularized and Orthogonal NMF \cite{Chen2024, Liu2025, Li2025, Wang2021} preserve local data structure or reduce component redundancy. While effective for clustering, these methods either alter small but relevant isotope contributions or impose constraints (e.g., orthogonality) that conflict with overlapping peaks in different isotope spectra. \textsc{gamma\_flow} instead uses a cosine similarity check for outlier detection while preserving all physically relevant peaks.

Subspace-based extensions \cite{Naseri2025, Zhou2020} learn low-dimensional manifolds or incorporate subspace priors to improve robustness. In contrast, \textsc{gamma\_flow} defines the latent space a priori from reference spectra, enabling supervised and physically interpretable unmixing.

Finally, supervised NMF approaches \cite{Leuschner2019, Lee2010, Bisot2016} often learn latent components and coefficients jointly via classification loss. Meanwhile, \textsc{gamma\_flow} fixes the basis from labelled training data and estimates coefficients via non-negative least squares, which reduces computational cost while ensuring physical interpretability.

\subsection{Limitations}

The applicability of \textsc{gamma\_flow} is limited by its \textbf{reliance on a predefined basis}, making it suitable only when relevant isotopes are known and labeled spectra are available. If only mixtures are present, an initial NMF decomposition is required to obtain training spectra. 

Furthermore, the \textbf{absence of orthogonality constraints} makes performance strongly dataset-dependent: similar isotope spectra can be confused, which may \textbf{limit scalability}. Preliminary tests successfully reduced 256 channels to 18 components, but the upper limit remains dataset-specific. Therefore, the correlation between latent components should be assessed based on their cosine similarity matrix (data exploration in \texttt{01\_preprocessing.ipynb}).

In addition, no intrinsic robustness exists against \textbf{mislabeled training spectra}, requiring careful expert validation. \textbf{Preprocessing choices}, especially rebinning parameters, can also affect results: parameters far outside the calibration range may cause data loss or resolution degradation, though this step can be skipped or adapted for other channel-energy relationships.

Broader \textbf{applicability to non-gamma spectra} is possible, but the fixed-basis approach assumes that loadings match the resolution and peak shapes of the target data. Substantially broader or asymmetric peaks may reduce reconstruction accuracy and classification performance, but this is mitigated by deriving the loadings from spectra acquired under comparable conditions. 

In theory, the non-negativity constraint only applies to the scores matrix, enabling \textsc{gamma\_flow} to analyze spectra containing both positive and negative signals, e.g. difference spectra. However, as those datasets are uncommon in most spectroscopic applications, the algorithm has not yet been tested for these use cases.

\section{Conclusions}
In this paper, we introduced \textsc{gamma\_flow}, an open-source, Python-based framework for the analysis of one-dimensional spectra. The approach combines established data science techniques in a resource-efficient way, enabling real-time analysis at low computational cost. Based on a supervised variant of non-negative matrix factorization, it constructs a fast and interpretable model for classification, decomposition, denoising, and outlier detection of spectral data. The method is applicable to both pure (single-label) and mixed (multi-label) spectra. 

For a user-friendly implementation, the software is organized into three jupyter notebooks. Each notebook combines explanatory text with code, guiding users step-by-step through data preprocessing and exploration, model training and evaluation, and outlier detection. This enables users not only to apply the model to spectral data but also develop a deep understanding of the underlying processes - without requiring advanced mathematical or programming expertise.

\textsc{gamma\_flow} bridges the gap between complex, often proprietary spectral analysis tools and the limitations of manual interpretation in laboratory workflows. By offering an open, transparent, and extensible framework, it empowers researchers to move beyond traditional black-box approaches. With its modular architecture and general applicability, it opens new avenues for innovation across disciplines — from nuclear safety and materials science to environmental monitoring and beyond. As such, \textsc{gamma\_flow} has the potential to accelerate scientific discovery wherever spectral data plays a role.

\section*{Acknowledgements}
We gratefully acknowledge the support provided by the Federal Ministry for the Environment, Climate Action, Nature Conservation and Nuclear Safety (BMUKN), whose funding has been instrumental in enabling us to achieve our research objectives and explore new directions. 
We also extend our appreciation to Martin Bussick in his function as the AI coordinator. 
Additionally, we thank the entire AI-Lab team for their support and inspiration, with special recognition to Ruth Brodte for guidance on legal and licensing matters.

\bibliographystyle{elsarticle-num} 
\bibliography{gamma_flow_neu.bib}

\end{document}